%% file: pmlr-sample.tex
\documentclass[pmlr]{jmlr}

\usepackage{longtable}

\usepackage{booktabs}
\usepackage[load-configurations=version-1]{siunitx} 

\theorembodyfont{\upshape}
\theoremheaderfont{\scshape}
\theorempostheader{:}
\theoremsep{\newline}

\usepackage{booktabs}
\usepackage{makecell}
\usepackage{multirow}

\jmlrvolume{}
\jmlrpages{}
\jmlrproceedings{JMLR}{}
\jmlrworkshop{}

\title[Linguistic Features for LLM-based AES]{Improve LLM-based Automatic Essay Scoring with Linguistic Features}

 \author{
    \Name{Zhaoyi Joey Hou} \Email{joey.hou@pitt.edu}\\
    \Name{Alejandro Ciuba} \Email{alejandrociuba@pitt.edu}\\
    \Name{Xiang Lorraine Li} \Email{xianglli@pitt.edu}\\
}

\begin{document}

\maketitle

\begin{abstract}
\input{sections/0-abstract}
\end{abstract}

\input{sections/1-intro}

\input{sections/2-Related-Works}

\input{sections/3-Methods}

\input{sections/4-Experiment}
\input{sections/5-Discussion}

\input{sections/6-Conclusion}

\input{sections/7-Limitations}

\input{sections/8-Acknowledgements}

\bibliography{citations/alejandro-citations, citations/joey-citations}

\appendix

\input{sections/app-ellipse-rubric}

\input{sections/app-supervised-baseline}

\input{sections/app-scoring-prompts}
\input{sections/app-linguistic-features}
\input{sections/app-parsing}

\end{document}

%% file: sections/0-abstract.tex

Automatic Essay Scoring (AES) assigns scores to student essays, reducing the grading workload for instructors. Developing a scoring system capable of handling essays across diverse prompts is challenging due to the flexibility and diverse nature of the writing task. Existing methods typically fall into two categories: supervised feature-based approaches and large language model (LLM)-based methods. Supervised feature-based approaches often achieve higher performance but require resource-intensive training. In contrast, LLM-based methods are computationally efficient during inference but tend to suffer from lower performance. This paper combines these approaches by incorporating linguistic features into LLM-based scoring. Experimental results show that this hybrid method outperforms baseline models for both in-domain and out-of-domain writing prompts\footnote{Codebase: \url{https://github.com/JoeyHou/essay_eval}}.


%% file: sections/1-intro.tex
\section{Introduction}
\label{introduction}
Research in Automatic Essay Scoring (AES), the task of automatically assessing the quality of an essay, dates back to over five decades ago \cite{page1968}. Since then, researchers in this domain have taken various perspectives; some focus on building hand-crafted features \cite{chen-he-2013-automated, uto-etal-2020-neural}, some leverage the computational power of neural-network to learn effective representation of essays \cite{dong-etal-2017-attention, ridley2020promptagnosticessayscorer, jin-etal-2018-tdnn}, and some adapt pre-trained language models as the starting point for fine-tuning \cite{wang-etal-2022-use, hierarchicalbert}. 

However, as much of the research above has shown, AES remains an open question. The key challenge of this task lies in the balance between generalization and specification: ideally, the method should be applicable to any grading scenario given a concrete grading rubric. Yet, numerous factors, including but not limited to instructors, education institutions, the essay's purpose, and the type of the essay (from the literature point of view) make essay grading context-specific. To this end, the cross-prompt AES system, which aims to work similarly well for different essay prompts and scoring rubrics, has been an important direction that draws much attention~\cite{ridley2020promptagnosticessayscorer, jin-etal-2018-tdnn, li-ng-2024-conundrums, phandi-etal-2015-flexible,Ridley_He_Dai_Huang_Chen_2021}.


 Creating cross-prompt AES systems with sufficient capabilities requires the system to examine more than just simple word surface forms and incorporate more linguistically motivated features (e.g., \cite{burstein1998}). Recent work (\cite{ridleyPromptAgnosticEssay2020, uto-etal-2020-neural}) pairs linguistic features with supervised methods to boost AES quality. However, there is little work on exploring linguistic features in the context of instruction-based large language models (LLMs), with most work focusing on other features such as rubric-incorporation \cite{hashemi-etal-2024-llm} and prompting techniques (e.g., \cite{liu-etal-2023-g} and \cite{chiang-lee-2023-closer}. To bridge the gap, we explore adding linguistic features to the LLM prompt. We conduct experiments in the cross-prompt AES setting with both open- (Mistral) and closed-source (GPT-4) large language models and find that LLMs align better with human judgments when given linguistic features. 
 
Our main contribution can be summarized as follows: 1) through prompt tuning and feature engineering, we have shown incorporating linguistic features into existing zero-shot prompting methods can notably improve the overall score prediction; 2) even for out-of-distribution data (i.e., essay from an entirely different dataset), the improvement holds; 3) there is still notable headroom for open-source LLM to automatically evaluate student essay, compared to their closed-source counterparts and smaller, supervised language models; we hypothesize that it is due to poorly-calibrated prior that is built-in to the LLM.

%% file: sections/2-Related-Works.tex
\section{Related Works}
\label{related-works}
\subsection{Automatic Essay Scoring}
\label{automatic-essay-scoring}

\paragraph{Feature Engineering} approaches leverage various features to predict essay scores, including linguistic features, e.g., readability metrics and word length ~\cite{ridley2020promptagnosticessayscorer, uto-etal-2020-neural, jin-etal-2018-tdnn, FoltLahaLand19993j, chen-he-2013-automated}, and content features, e.g., content quality and organization ~\cite{mathias-bhattacharyya-2018-asap, crossley2023english}. Models that utilize these features range from simple logistic regression models ~\cite{chen-he-2013-automated} to deep neural networks ~\cite{uto-etal-2020-neural}. These approaches assess the quality of essays in an interpretable manner with well-defined features.

\paragraph{Language-model-based} approaches emerge with the rising popularity of Transformer architecture, including BERT-based methods that require supervised fine-tuning \cite{wang-etal-2022-use, mutlitaskAESforEssayGrading, hierarchicalbert} and LLM-based methods that focus on prompt-engineering \cite{mansour-etal-2024-large, stahl-etal-2024-exploring}. In particular, \cite{stahl-etal-2024-exploring} explores zero-shot prompting with persona prompts and analysis instructions. Building on this, our work aims to utilize linguistic features in LLM prompting.

\subsection{LLM as Evaluator}
\label{llm-as-evaluator}
Given the increasing capability of LLMs and their scalable nature, researchers in various domains have explored how to use them for the automatic evaluation of text content \cite{zubiaga-etal-2024-llm, alhafni-etal-2024-personalized, gao2024llmbasednlgevaluationcurrent, fu-etal-2024-gptscore}. Although some research has shown proper prompt tuning, such as explanation-guided generation, clear rubric guidance, and chain-of-thought (COT) could improve the alignment between human and LLMs \cite{chiang-lee-2023-closer, liu-etal-2023-g, hashemi-etal-2024-llm}, the LLM-based evaluators still perform underwhelming in more complex tasks, such as reviewing papers \cite{zhou-etal-2024-llm} and scoring students essay \cite{mansour-etal-2024-large, stahl-etal-2024-exploring}. In this work, we specifically focus on improving LLMs as student essay graders by incorporating the linguistic features of essays. Additionally, we examine the transferability of the prompts, i.e., how a prompt that is tuned in the in-distribution data would perform out-of-distribution in the same task.

%% file: sections/3-Methods.tex
\section{Methods}
\label{methods}
\subsection{Zero-shot Prompts with LLM}
We build on top of the prompt template and instruction strategy by \cite{stahl-etal-2024-exploring} for our zero-shot prompt design. Each prompt follows the following structure: \textbf{persona pattern}, \textbf{essay prompt}, \textbf{analysis task}, \textbf{student's essay}, \textbf{additional information}, and \textbf{format instruction}. We adopt the best-performing combination of each component based on \cite{stahl-etal-2024-exploring}, i.e., \textit{educational researcher} as the prompt template, \textit{Explanation → Scoring} as the analysis instruction. We also add the \textbf{additional information} section to incorporate linguistic features of the essay (see Sec. \ref{linguistic-features-2}). Below is the prompt template we use. More details about the prompt structure and examples can be found in Appendix \ref{app-scoring-prompts}.
\begin{center}
\footnotesize
\begin{quote}
You are part of an educational research team analyzing the writing skills of students in grades 7 to 10. You have been given a student's essay and the prompt they responded to.

\#\#\# Essay Prompt: \textit{\{ essay prompt \}}

\#\#\# Analysis Task: \textit{\{ analysis instruction \}}

\#\#\# Analyzed Student Essay: \textit{\{ essay \}}

\#\#\# Additional Information: Studies show that the following features are highly, positively correlated with the grade of the essay (i.e., higher features typically mean higher end score): \textit{\{ linguistic features \}}

\#\#\# Analysis: Conclude your analysis with a grade and comments in the following format: \textit{\{ format instruction \}}
\end{quote}
\end{center}

\subsection{Linguistic Features}
\label{linguistic-features-2}
In addition to the naive zero-shot baseline, we experimented with incorporating linguistic features into the prompts. We base our model's linguistic features off of \cite{ridleyPromptAgnosticEssay2020}, which were additionally used by \cite{li-ng-2024-conundrums} and cited as being some of the most impactful features in essay grading, having a Pearson's correlation score of 0.6 or above with the essay's score. While the original works contain much more linguistic features, limiting our linguistic features helped avoid overly long prompts, which could negatively impact LLM performance. The linguistic features we used are listed below.

\textbf{Unique Words} refers to the number of single-instance words in the essay. For \textbf{essay character length}, we only count the number of non-space, non-punctuation characters\footnote{Words are not normalized as in the original paper.}. \textbf{Word/Sentence counts} are the total number of words/sentences in a given essay. We get the \textbf{individual counts for the total number of lemmas, nouns, and stop-words.} Finally, we get the essay's \textbf{Dale-Chall} (\cite{dale1948}) word count, \textbf{total character count} (all characters) and \textbf{long word count.}

During prompt construction, the linguistic features are formatted as an unordered list of a short feature description followed by feature value. The formatted text containing all linguistic features is inserted into the prompt as the \textbf{additional information} section. 

\subsection{Output Parsing via LLM}
The output format varies across essay sets since each essay set has its own scoring schema. To make the pipeline generic to any input, we implement a few-shot parsing module powered by a stand-alone LLM. More details can be found in Appendix \ref{app-parsing}.

%% file: sections/4-Experiment.tex
\section{Experiments}
\label{experiments}
\subsection{Datasets \& Linguistic Features}
\label{dataset-and-evaluation-metrics}
We conducted our experiments on two widely used essay-grading datasets:

\paragraph{ASAP}
\label{asap-and-asap++}
ASAP (\cite{asap-aes}) is one of the most widely used evaluation datasets for AES, with 12,980 essays written by students in grades 7 through 10. It is divided into 8 subsets based on the formulation of essay prompts, including argumentative (1, 2), source-dependent (3, 4, 5, 6), and narrative (7, 8) essay prompts. The scores are annotated by at least two graders to ensure validity. We split the ASAP dataset into 5 equally sized folds. We use 3 folds for training (BERT baseline fine-tuning), 1 fold for validation (BERT baseline model selection and prompt tuning for prompting method), and 1 fold for evaluation.

\paragraph{ELLIPSE}
\label{ellipse}
\textit{English Language Learners Insight, Proficiency and Skills Evaluation} (\textit{ELLIPSE}) by \cite{crossleyEnglishLanguageLearner2024} is composed of 6483 essays from English language learners in the United States educational system's grades 8-12 get from an unnamed standardized test. The dataset features the full essay of each student and demographic information such as race/ethnicity and income background. There are 29 different argumentative essay prompts covering topics such as cell phones in school and community service. Each essay includes an overall holistic score as well as six rubric-based scores, all scored from one to five (in intervals of 0.5) by one of two annotators (see Appendix \ref{ellipse-rubric} for the full rubric). Note that we do not split the data since we want to treat this dataset as an entirely out-of-distribution dataset to examine the generalizability of our method across different essay prompts and contents.

\paragraph{Linguistic Features}
\label{linguistic-features}

While the ASAP dataset had these features pre-calculated thanks to the aforementioned prior work, the ELLIPSE dataset needed a custom pipeline to re-implement these features. We followed \cite{ridleyAutomatedCrosspromptScoring2021} as exactly as possible to recreate the features from Section \ref{linguistic-features-2} as closely as possible. See Appendix \ref{app:linguistic-features} for the implementation details.

\subsection{Evaluation Metrics}
For both ASAP and ELLIPSE, the most important task is to predict the overall scoring. We follow the Kaggle competition evaluation metric of \cite{asap-aes} and use Quadratic Weighted Kappa (QWK) to measure the agreement between the predicted and annotated overall scores.


\subsection{Models and Experiment Setup}
\label{models-and-experiment-setup}

\paragraph{Our Pipeline}
\label{our-pipeline-(Mistral7B)}
As mentioned in Sec. \ref{methods}, our main method consists of a prompt construction module and a zero-shot prompting module with an LLM followed by a parsing module powered by another LLM. For prompt construction, we include three setups to integrate linguistic features: the most correlated feature (i.e., unique word count), the top 3 correlated features (i.e., unique word count, lemma count, and complex word count), and all 10 features, described in Sec. \ref{linguistic-features}. For prompting and parsing, we use \texttt{Mistral-7B-Instruct-v0.2} (\cite{jiang2023mistral7b}) for all LLM-related jobs, and we use 0 temperature and 4096 max token length, following \cite{stahl-etal-2024-exploring}\footnote{In early experiments, Llama 3 results are significantly worse than others; so we drop it and keep Mistral.}. We use default sampling parameters in \texttt{vllm} framework during decoding\footnote{\url{https://docs.vllm.ai/en/v0.6.1.post1/dev/sampling_params.html}}. All experiments are conducted with the VLLM\footnote{\url{https://docs.vllm.ai/en/v0.6.1.post1/index.html}} (\cite{kwon2023efficient}) on a single-card NVIDIA L40S device. 

\paragraph{Unsupervised Baseline (GPT-4)}
\label{unsupervised-baseline-(gpt4)}
We also include one of the most capable models available to us, GPT-4 (\texttt{gpt-4-0613}), as the strong, unsupervised baseline. The experiment setup is exactly the same as in our pipeline above, with two differences, both due to the limited budget: 1) we only experiment with no more than 500 randomly sampled essays per essay set in ASAP (some essay sets have less than 500 essays in the test set) and 500 randomly sampled essays in ELLIPSE; 2) we only experiment with no linguistic feature and the best-performing linguistic feature setup (i.e., all top 10 features) based on the performance on the \texttt{dev} set of ASAP.

\paragraph{Supervised Baseline (BERT)}
\label{supervised-baseline-(bert)}
We also experimented a supervised method on ASAP to establish a performance upper bound. Our supervised baseline is a BERT-based architecture utilizing three main feature classes: document-, token- and segment-scale features~\cite{wang-etal-2022-use}. We base our fine-tuning on the authors' available code. The authors only released the fine-tuned model for ASAP prompt 8, so---to obtain models for all prompts---we fine-tuned \texttt{bert-base-uncased} as they did in the original paper. We used our splits of the ASAP dataset for fine-tuning, validation and testing (see \ref{asap-and-asap++}). More details about fine-tuning is in Appendix \ref{supervised-baseline-details}.

%% file: sections/5-Discussion.tex
\section{Result and Discussion}
\label{discussion}

\subsection{Cross-dataset Performances}
\label{cross-dataset-performances}
The experimental results are summarized in Table~\ref{table:results}. For ASAP, the BERT-based supervised baseline achieved the highest average QWK, aligning with our expectation that AES remains a challenging task for naive LLMs prompting. When comparing cross LLMs, the performance between GPT-4 and Mistral showed mixed results across the two datasets.
We suspect this is due to the differences in the essay type. ELLIPSE essays are argumentative, similar to essay sets 1 and 2 in ASAP. However, the grading rubric for ASAP essay set 2 is more specific (focusing on "Writing Application" and "Language Conventions") compared to the broader "Overall" quality criteria used for ELLIPSE and ASAP essay set 1. These observations suggest that LLM performance on ELLIPSE is more comparable to ASAP essay set 1 rather than the entire ASAP dataset, a pattern confirmed by the results in Table~\ref{table:results}. Another research \cite{mutlitaskAESforEssayGrading} that conducted a zero-shot prompting experiment on ELLIPSE with ChatGPT \footnote{\url{https://chatgpt.com/}} also got similar results (QWK = 0.29).

\subsection{Benefit of Linguistic Features}
\label{effect-of-linguistic-features}
As shown in Table \ref{table:results}, the prompts with linguistic features almost always perform better than the ones without -- with the exception of GPT-4. This trend holds even for out-of-distribution data (ELLIPSE). When it comes to Mistral 7B, for ASAP, \textit{Top-10} features are the most effective by both average and subset QWK measurement; for ELLIPSE, the best performing linguistic feature choice is only \textit{Top-3} feature, while the difference is marginal. In addition, the improvement brought by the linguistic features even pushes the Mistral performance on ASAP close to GPT-4 performance. Based on these observations, we can conclude that including linguistic features can benefit LLM-based zero-shot AES.

\begin{table*}
\small
\setlength{\tabcolsep}{3pt}
\centering
  \begin{tabular}{cc|ccccccccc|c}
    \toprule
    \multirow{2}{*}{\textbf{Model}} &
    \multirow{2}{*}{\textbf{\shortstack{Linguistic \\ Features}}} &
      \multicolumn{9}{c}{\textbf{ASAP}} & 
     \multirow{2}{*}{\textbf{ELLIPSE}} \\
        & & \textbf{Avg.} & 1 & 2 & 3 & 4 & 5 & 6 & 7 & 8 & \\
    \midrule
        \multirow{1}{*}{BERT}
            & None  & .545 & .741 & .447  & .331  & .430 & .734  & .552 & .715 & .413 & N/A  \\
    
    \midrule
        \multirow{2}{*}{\shortstack{GPT-4}}
            & None  & \textbf{.499} & .221 & .581  & \textbf{.514}  & \textbf{.631} & .561  & \textbf{.686} & .250 & \textbf{.553} & .307  \\
            & Top 10  & .488 & \textbf{.285} & \textbf{.592} & .444 & .578 &  \textbf{.620} & .645 & \textbf{.251} & .491 &  \textbf{.345}  \\

    \midrule     		
        \multirow{4}{*}{Mistral 7B}
            & None & .454 & .254 & .474 & \textbf{.526} & .549 &  .506 & \textbf{.567} & .388 & \textbf{.367} & .454  \\
            & Unique Word & .458 & .362 & \textbf{.516} & .454 & .552 &  .492 & .539 & .438 & .313 & .475  \\
            & Top 3  & .461 & .383 & .516 & .453 & .567 & .503 & .542 & .409 & .318 & \textbf{.481} \\
            & Top 10  & \textbf{.492} & \textbf{.423} & .483 & .493 & \textbf{.623} &  \textbf{.511} & .537 & \textbf{.508} & .360 & .468  \\ 
    \toprule
  \end{tabular}
  
  \caption{Results of our experiments; BERT model is trained on ASAP training set and not applicable to ELLIPSE; Bold numbers are best-performing setups of the model.}
  \label{table:results}
\end{table*}

%% file: sections/6-Conclusion.tex
\section{Conclusion}
\label{conclusion}
In this work, we explore the combination of linguistic features and zero-shot prompting with SoTA LLMs in the task of automatic essay scoring. Empirical experiments show performance improvement when linguistic features are integrated into the zero-shot prompt. However, the performance improvement varies depending on the type of the essay, proving the challenging nature of generalizability in AES systems. We hope our work can serve as a starting point for the research in more interpretable and more generalizable LLM-based AES methods.


%% file: sections/7-Limitations.tex
\section{Limitations}
\label{limitations}

During our experiments, we noted several limitations that future work could expand upon and resolve. Firstly, there is only one open-source and one close-source one. Future work could look at including more LLMs for comparison. Secondly, the prediction target, holistic score, is still hard to interpret, whereas some subsets (7 and 8) of ASAP and the entire set of ELLIPSE do have fine-grained essay score annotations (see examples in Appendix \ref{ellipse-rubric}). Incorporating them into the overall score prediction process would make the overall score more transparent. Thirdly, the persona section of the prompt template mentions ``grade 7 to 10,'' which is the age range for students in ASAP; however, the students in the ELLIPSE dataset are from grades 8 to 12, which might lead to performance differences among those two datasets. Lastly, our datasets have a clear western bias, especially ELLIPSE, which focuses on ESL students in the United States. We believe the community would benefit from more diverse and inclusive datasets.

%% file: sections/8-Acknowledgements.tex

%% file: sections/app-ellipse-rubric.tex
\pagebreak
\section{ELLIPSE Rubric}
\label{ellipse-rubric}
\begin{center}
    \small
    \begin{longtable}{|p{0.12\linewidth}|p{0.15\linewidth}|p{0.15\linewidth}|p{0.15\linewidth}|p{0.15\linewidth}|p{0.15\linewidth}|}
        \hline
        Score \ \ \ \ Category & 5                                                                                                                                                                                                                                                & 4                                                                                                                                                                                                                       & 3                                                                                                                                                                                                                                                                         & 2                                                                                                                                                                                                               & 1                                                                                                                                                                                            \\ \hline
        Overall                       & Native-like facility in the use of language with syntactic variety, Appropriate word choice and phrases; well-controlled text organization; precise use of grammar and conventions; rare language inaccuracies that do not impede communication. & Facility in the use of language with syntactic variety and range of words and phrases; controlled organization; accuracy in grammar and conventions; occasional language inaccuracies that rarely impede communication. & Facility limited to the use of common structures and generic vocabulary; organization generally controlled although connection sometimes absent or unsuccessful; errors in grammar and syntax and usage. Communication is impeded by language inaccuracies in some cases. & Inconsistent facility in sentence formation, word choice, and mechanics; organization partially developed but may be missing or unsuccessful. Communication impeded in many instances by language inaccuracies. & A limited range of familiar words or phrases loosely strung together; frequent errors in grammar (including syntax) and usage. Communication impeded in most cases by language inaccuracies. \\ \hline
        Cohesion                      & Text organization consistently well controlled using a variety of effective linguistic features such as reference and transitional words and phrases to connect ideas across sentences and paragraphs; appropriate overlap of ideas.             & Organization generally well controlled; a range of cohesive devices used appropriately such as reference and transitional words and phrases to connect ideas; generally appropriate overlap of ideas                    & Organization generally controlled; cohesive devices used but limited in type; Some repetitive, mechanical, or faulty use of cohesion use within and/or between sentences and paragraphs.                                                                                  & Organization only partially developed with a lack of logical sequencing of ideas; some basic cohesive devices used but with inaccuracy or repetition.                                                           & No clear control of organization; cohesive devices not present or unsuccessfully used; presentation of ideas unclear.                                                                        \\ \hline
        Syntax                        & Flexible and effective use of a full range of syntactic structures including simple, compound, and complex sentences; There may be rare minor and negligible errors in sentence formation.                                                       & Appropriate use of a variety of syntactic structures, such as simple, compound, and complex sentences; occasional errors or inappropriateness in sentence formation.                                                    & Simple, compound, and complex syntactic structures present although the range may be limited; some apparent errors in sentence formation, especially in more complex sentences.                                                                                           & Some sentence variation used; many sentence structure problems.                                                                                                                                                 & Pervasive and basic errors in sentence structure and word order that cause confusion; basic sentences errors common.                                                                         \\ \hline
        Vocabulary                    & Wide range of vocabulary flexibly and effectively used to convey precise meanings; skillful use of topic-related terms and less common words; rare negligible inaccuracies in word use.                                                          & Sufficient range of vocabulary to allow flexibility and precision; appropriate use of topic-related terms and less common lexical items                                                                                 & Minimally adequate range of vocabulary for the topic; no precise use of subtle word meanings; topic related terms only used occasionally; attempts to use less common vocabulary but with some inaccuracy                                                                 & Narrow range of vocabulary to convey basic and elementary meaning; topic related terms used inappropri ately; errors in word formation and word choice that may distort meanings                                & Limited vocabulary often inappropriately used; limited control of word choice and word forms; little attempt to use topic-related terms                                                      \\ \hline
        Phraseology                   & Flexible and effective use of a variety of phrases, such as idioms, collocations, and lexical bundles, to convey precise and subtle meanings; rare minor inaccuracies that are negligible.                                                       & Appropriate use of a variety of phrases, such as idioms, collocations, and lexical bundles; occasional inaccuracies and colloquialisms.                                                                                 & Evident use of phrases such as idioms, collocations, and lexical bundles but without much variety; some noticeable repetitions and misuses.                                                                                                                               & Narrow range of phrases, such as collocations and lexical bundles, used to convey basic and elementary meaning; many repetitions and /or misuses of phrases.                                                    & Memorized chunks of language, or simple phrasal patterns predominate; many repetitions and misuses of phrases.                                                                               \\ \hline
        Grammar                       & Command of grammar and usage with few or no errors.                                                                                                                                                                                              & Minimal errors in grammar and usage.                                                                                                                                                                                    & Some errors in grammar and usage.                                                                                                                                                                                                                                         & Many errors in grammar and usage.                                                                                                                                                                               & Errors in grammar and usage throughout.                                                                                                                                                      \\ \hline
        Conventions                   & Consistent use of appropriate conventions to convey meaning; spelling, capitalization, and punctuation errors nonexistent or negligible.                                                                                                         & Generally consistent use of appropriate conventions to convey meaning; spelling, capitalizatio n, and punctuation errors few and not distracting.                                                                       & Developing use of conventions to convey meaning; errors in spelling, capitalization, and punctuation that are sometimes distracting.                                                                                                                                      & Variable use of conventions; spelling, capitalization, and punctuation errors frequent and distracting.                                                                                                         & Minimal use of conventions; spelling, capitalizatio n, and punctuation errors throughout.                                                                                                    \\ \hline
    \caption{The ELLIPSE rubric, gotten directly from the original paper.}
    \label{tab:ellipse-rubric}
    \end{longtable}
\end{center}

%% file: sections/app-supervised-baseline.tex
\section{Supervised Baseline Details}
\label{supervised-baseline-details}
Our supervised baseline~\cite{wang-etal-2022-use}\footnote{\url{https://github.com/lingochamp/Multi-Scale-BERT-AES}}is a BERT-based architecture comprised of two sub-components---each pretrained BERT models (\cite{devlin})---which analyze three main feature classes: document-, token- and segment-scale features. The first sub-component receives the document- and token-scale features. It is fine-tuned to learn the document-scale feature representation through the \texttt{[CLS]} (start) token\footnote{There can be multiple text segments per essay as their input length is set to 510.} and the token-scale features through the BERT word embeddings. Its output goes through a final max pooling layer to represent the sub-component's score. The segment-scale features are received by the second sub-component, which takes in an essay as a series of segments each of size $k$ (except the last segment, which is smaller). A list of these segment series of varying sizes $k_{i}$ are input into the model sequentially, and a final LSTM and attention and dense pooling layer is used to output the sub-component's score. Lastly, the output from the two sub-components are added together to produce the final holistic score. The model's loss function is additive between mean squared error ($MSE$), cosine similarity ($CS$) and margin ranking loss ($MLR$): $\mathcal{L}_{\mathrm{Total}}(\mathbf{x}, \mathbf{y}) = \alpha \mathcal{L}_{MSE}(\mathbf{x}, \mathbf{y}) + \beta \mathcal{L}_{CS}(\mathbf{x}, \mathbf{y}) + \gamma \mathcal{L}_{MLR}(\mathbf{x}, \mathbf{y})$.

We base our fine-tuning on the authors' available code\footnote{Available upon acceptance.}. The authors only released their fine-tuned model for ASAP prompt 8, so---to obtain models for all prompts---we fine-tuned \texttt{bert-base-uncased} as they did in the original paper. We used our splits of the ASAP dataset for fine-tuning, validation and testing (see \ref{asap-and-asap++}). We fine-tune for 80 epochs, our hyperparameters for $\alpha$, $\beta$ and $\gamma$ were all set to 0.5 and with cosine similarity \texttt{dim=1} and margin ranking loss \texttt{margin=0}. Everything is implemented in PyTorch (\cite{paszke2019}) and HuggingFace (\cite{wolf2019}) using \texttt{google-bert/bert-base-uncased}. We run the test set on the prompt's model with the best loss.

%% file: sections/app-scoring-prompts.tex
\section{Zero-shot Essay Scoring Prompts}
\label{app-scoring-prompts}
Here are some examples of zero-shot essay scoring prompts. Note that the exact phrasing and wording are not exactly the same as \cite{stahl-etal-2024-exploring} paper. That is because we have failed to reproduce the exact same results in their paper, motivating us to conduct a limited prompt tuning in the \texttt{dev} set of ASAP. To reduce complexity, the tuning is done only in phrasing and formatting, without changing the overall structure of the prompt compared to the original design.

\subsubsection{No Linguistic Feature}
\textit{
You are part of an educational research team analyzing the writing skills of students in grades 7 to 10. You have been given a student's essay and the prompt they responded to.  \\
\#\#\# Essay Prompt: More and more people use computers, but not everyone agrees that this benefits society. Those who support advances in technology believe that computers have a positive effect on people. They teach hand-eye coordination, give people the ability to learn about faraway places and people, and even allow people to talk online with other people. Others have different ideas. Some experts are concerned that people are spending too much time on their computers and less time exercising, enjoying nature, and interacting with family and friends. Write a letter to your local newspaper in which you state your opinion on the effects computers have on people. Persuade the readers to agree with you.  \\
}
\textit{
\#\#\# Analysis Task: Grade the given essay with the following requirements:  \\
- Use those score ranges: Overall: from 1 to 6.  \\
- Provide an explanation for your score as well.  \\
}
\textit{
\#\#\# Analyzed Student Essay: Dear, @CAPS1 @CAPS2 @CAPS3 More and more people use computers, but not everyone agrees that this benefits society. Those who support advances in technology believe that computers have a positive effect on people. Others have different ideas. A great amount in the world today are using computers, some for work and spme for the fun of it. Computers is one of mans greatest accomplishments. Computers are helpful in so many ways, @CAPS4, news, and live streams. Don't get me wrong way to much people spend time on the computer and they should be out interacting with others but who are we to tell them what to do. When I grow up I want to be a author or a journalist and I know for a fact that both of those jobs involve lots of time on time on the computer, one @MONTH1 spend more time then the other but you know exactly what @CAPS5 getting at. So what if some expert think people are spending to much time on the computer and not exercising, enjoying natures and interacting with family and friends. For all the expert knows that its how must people make a living and we don't know why people choose to use the computer for a great amount of time and to be honest it's non of my concern and it shouldn't be the so called experts concern. People interact a thousand times a day on the computers. Computers keep lots of kids of the streets instead of being out and causing trouble. Computers helps the @ORGANIZATION1 locate most wanted criminals. As you can see computers are more useful to society then you think, computers benefit society. \\
}
\textit{
\#\#\# Analysis: Conclude your analysis with a grade and comments in the following format:  \\
\#\#\# Explanation:  \\
\#\#\# Score: \\
- Overall:
}

\subsubsection{Top-10 Features}
\textit{
You are part of an educational research team analyzing the writing skills of students in grades 7 to 10. You have been given a student's essay and the prompt they responded to.  \\
\#\#\# Essay Prompt: More and more people use computers, but not everyone agrees that this benefits society. Those who support advances in technology believe that computers have a positive effect on people. They teach hand-eye coordination, give people the ability to learn about faraway places and people, and even allow people to talk online with other people. Others have different ideas. Some experts are concerned that people are spending too much time on their computers and less time exercising, enjoying nature, and interacting with family and friends. Write a letter to your local newspaper in which you state your opinion on the effects computers have on people. Persuade the readers to agree with you.  \\
}
\textit{
\#\#\# Analysis Task: Grade the given essay with the following requirements:  \\
- Use those score ranges: Overall: from 1 to 6.  \\
- Provide an explanation for your score as well.  \\
}
\textit{
\#\#\# Analyzed Student Essay: Dear, @CAPS1 @CAPS2 @CAPS3 More and more people use computers, but not everyone agrees that this benefits society. Those who support advances in technology believe that computers have a positive effect on people. Others have different ideas. A great amount in the world today are using computers, some for work and spme for the fun of it. Computers is one of mans greatest accomplishments. Computers are helpful in so many ways, @CAPS4, news, and live streams. Don't get me wrong way to much people spend time on the computer and they should be out interacting with others but who are we to tell them what to do. When I grow up I want to be a author or a journalist and I know for a fact that both of those jobs involve lots of time on time on the computer, one @MONTH1 spend more time then the other but you know exactly what @CAPS5 getting at. So what if some expert think people are spending to much time on the computer and not exercising, enjoying natures and interacting with family and friends. For all the expert knows that its how must people make a living and we don't know why people choose to use the computer for a great amount of time and to be honest it's non of my concern and it shouldn't be the so called experts concern. People interact a thousand times a day on the computers. Computers keep lots of kids of the streets instead of being out and causing trouble. Computers helps the @ORGANIZATION1 locate most wanted criminals. As you can see computers are more useful to society then you think, computers benefit society. \\
}
\textit{
\#\#\# Additional Information: Studies show that the following features are highly, positively correlated with the grade of the essay (i.e., higher features typically means higher end score)  \\
- total number of unique words in the essay: 113  \\
- total number of words in the essay.: 279  \\
- total number of sentences present: 14  \\
- total number of characters: 279  \\
- total number of lemma: 133  \\
- total number of nouns: 50  \\
- total number of stopwords: 71  \\
- total number of words that are not in the Dale-Chall word list of 3000 words recognized by 80\% of fifth graders: 80  \\
- total number of characters: 1229  \\
}
\textit{
\#\#\# Analysis: Conclude your analysis with a grade and comments in the following format:  \\
\#\#\# Explanation:  \\
\#\#\# Score: \\
- Overall:
}

%% file: sections/app-linguistic-features.tex
\section{Linguistic Features}
\label{app:linguistic-features}
\textbf{Unique words} refers to the number of single-instance words in the essay. For \textbf{essay character length}, we only count the number of non-space, non-punctuation characters. Words are not normalized before these metrics as in the original paper. \textbf{Total word count} and \textbf{total sentence count} per essay are gotten via \texttt{nltk} (\cite{loper2002}) tokenizers. We additionally utilize the \texttt{en\_core\_web\_sm} in \texttt{spaCy} (\cite{honnibal2020}) to get \textbf{separate counts for lemma, noun, and stop-words.} Finally, we get \textbf{the Dale-Chall} (\cite{dale1948}) \textbf{word count}, \textbf{total character count} and \textbf{long word count} with the \texttt{readability}\footnote{\url{https://pypi.org/project/readability/}} Python package.

Our implementation is based on the code \footnote{\url{https://github.com/robert1ridley/cross-prompt-trait-scoring/blob/main/features.py}} from the original paper \cite{ridleyAutomatedCrosspromptScoring2021}. Our implementation will be made available upon acceptance.

%% file: sections/app-parsing.tex
\section{Parsing Module}
\label{app-parsing}
\subsection{Configurations}
\begin{itemize}
    \item Model: Mistral-7B (the same configuration as the scoring model)
    \item Overall parsing error is less than 7\%.
\end{itemize}

\subsection{Few-shot Output Parsing}
\textit{
    You are an AI agent that specialized in converting text input into JSON format.\\
    Instruction: \\
    - Input: text with one or more score and some other relevant information (e.g., explanation, feedbacks, etc.)\\
    - Output: JSON text with `Score' as a mandatory key and other information organized by their field names\\
    - Make sure ONLY return the VALID JSON data, without any additional text or characters.\\
    Here are some examples
}\\

\textit{
Example Input:\\
\#\#\# Explanation: The student's essay demonstrates a limited understanding of the topic and a lack of cohesion. The essay jumps from one idea to another without a clear connection between them. The writing is also filled with numerous grammatical errors, misspellings, and inconsistent capitalization. \\
\#\#\# Score:\\
- Overall: 1 The essay demonstrates a very limited understanding of the topic and contains numerous errors in grammar, spelling, and capitalization. The writing lacks cohesion and a clear thesis statement, and the arguments are not well-supported with evidence or examples. \\
Example Output:\\
\{
    ``Score'': \{
        ``Overall'': 1
    \},
    ``Explanation'': ``The student's essay demonstrates a limited understanding of the topic and a lack of cohesion. The essay jumps from one idea to another without a clear connection between them. The writing is also filled with numerous grammatical errors, misspellings, and inconsistent capitalization.''
\}
}\\

\textit{
Example Input:\\
\#\#\# Explanation: The student's essay demonstrates a basic understanding of the topic and presents a clear position, but the writing is disorganized and contains numerous errors in language conventions. The essay jumps between discussing censorship in libraries and specific examples of offensive music, making it difficult to follow the main argument. \\
\#\#\# Score: \\
- Writing Applications: 2 The essay presents a viewpoint on the issue of censorship, but the argument is not well-developed or clearly stated. The student uses personal experiences and examples. 
- Language Conventions: 1 The essay contains numerous errors in language conventions, including incorrect capitalization, punctuation, and sentence structure. \\
Example Output:\\
\{
    ``Score'': \{
        ``Writing Applications'': 2,
        ``Language Conventions'': 1
    \}
    ``Explanation'': ``The student's essay demonstrates a basic understanding of the topic and presents a clear position, but the writing is disorganized and contains numerous errors in language conventions. The essay jumps between discussing censorship in libraries and specific examples of offensive music, making it difficult to follow the main argument.''
\} 
}\\

\textit{
Example Input:\\
\#\#\# Explanation: The student's essay demonstrates a moderate level of awareness of the audience, as they address the readers directly and use a conversational tone. \\
\#\#\# Feedbacks: the essay could have been more effective if the student had used more formal language and addressed specific concerns of the local community regarding the overuse of computers. \\
\#\#\# Score: \\
- Overall: 3 The student's essay shows some awareness of the audience, but there is room for improvement in terms of language and organization. The essay could benefit from more specific examples and a clearer, more focused argument. \\
Example Output:
\{
    ``Score'': \{
        ``Overall'': 3
    \},
    ``Explanation'': ``The student's essay demonstrates a moderate level of awareness of the audience, as they address the readers directly and use a conversational tone.'',
    ``Feedbacks'': "the essay could have been more effective if the student had used more formal language and addressed specific concerns of the local community regarding the overuse of computers.''
\} 
}\\
\textit{
Now work on the following input:\\
Input:\\
\{LLM OUTPUT\} \\
Output:
}